\documentclass[10pt,twocolumn,letterpaper]{article}

\usepackage{iccv}
\usepackage{times}
\usepackage{epsfig}
\usepackage{graphicx}
\usepackage{subfigure}
\usepackage{amsmath}
\usepackage{amssymb}
\usepackage{multirow}
\usepackage{stfloats}
\usepackage{tabularx}

\usepackage[pagebackref=true,breaklinks=true,colorlinks,bookmarks=false]{hyperref}

\newcommand\blfootnote[1]{%
  \begingroup
  \renewcommand\thefootnote{}\footnote{#1}%
  \addtocounter{footnote}{-1}%
  \endgroup
}

\iccvfinalcopy % *** Uncomment this line for the final submission

\def\iccvPaperID{*} % *** Enter the ICCV Paper ID here

\ificcvfinal\fi
\begin{document}
%end

\newcommand{\todo}[1]{{\textcolor{red}{TODO: #1}}}
\newcommand{\ct}[1]{\fontsize{7pt}{1pt}\selectfont{#1}}
\newcolumntype{x}{>\small c}
\definecolor{mygreen}{rgb}{0,0.5,0}

%%%%%%%%% TITLE
\title{Improving Object Detection With One Line of Code}

\author{Navaneeth Bodla* \hspace{20pt} Bharat Singh* \hspace{20pt} Rama Chellappa \hspace{20pt} Larry S. Davis\\
Center For Automation Research, University of Maryland, College Park\\
{\tt\small {\{nbodla,bharat,rama,lsd\}}@umiacs.umd.edu}
}

\maketitle

\blfootnote{*The first two authors contributed equally to this paper.}

%%%%%%%%% ABSTRACT
\begin{abstract}
Non-maximum suppression is an integral part of the object detection pipeline. First, it sorts all detection boxes on the basis of their scores. The detection box $\mathcal{M}$ with the maximum score is selected and all other detection boxes with a significant overlap (using a pre-defined threshold) with $\mathcal{M}$ are suppressed. This process is recursively applied on the remaining boxes. As per the design of the algorithm, if an object lies within the predefined overlap threshold, it leads to a miss. To this end, we propose Soft-NMS, an algorithm which decays the detection scores of all other objects as a continuous function of their overlap with $\mathcal{M}$. Hence, no object is eliminated in this process. Soft-NMS obtains consistent improvements for the coco-style mAP metric on standard datasets like PASCAL VOC 2007 (1.7\% for both R-FCN and Faster-RCNN) and MS-COCO (1.3\% for R-FCN and 1.1\% for Faster-RCNN) by just changing the NMS algorithm without any additional hyper-parameters. Using Deformable-RFCN, Soft-NMS improves state-of-the-art in object detection from 39.8\% to 40.9\% with a single model. Further, the computational complexity of Soft-NMS is the same as traditional NMS and hence it can be efficiently implemented. Since Soft-NMS does not require any extra training and is simple to implement, it can be easily integrated into any object detection pipeline. Code for Soft-NMS is publicly available on GitHub \url{http://bit.ly/2nJLNMu}.
\end{abstract}

%%%%%%%%% BODY TEXT
\section{Introduction}
%describe what is object detection

Object detection is a fundamental  problem in computer vision in which an algorithm generates bounding boxes for specified object categories and assigns them classification scores. It has many practical applications in autonomous driving \cite{dollar2009pedestrian,geiger2013vision}, video/image indexing \cite{sivic2003video,philbin2007object}, surveillance \cite{collins2000system,haritaoglu2000w} etc. Hence, any new component proposed for the object detection pipeline should not create a computational bottleneck, otherwise it will be conveniently ``ignored' in practical implementations. Moreover, if a complex module is introduced which requires re-training of models which leads to a little improvement in performance, it will also be ignored. However, if a simple module can improve performance without requiring any re-training of existing models, it would be widely adopted. To this end, we present a soft non-maximum suppression algorithm, as an alternative to the traditional NMS algorithm in the current object detection pipeline.

\begin{figure}
\centering
   \includegraphics[width=1\linewidth]{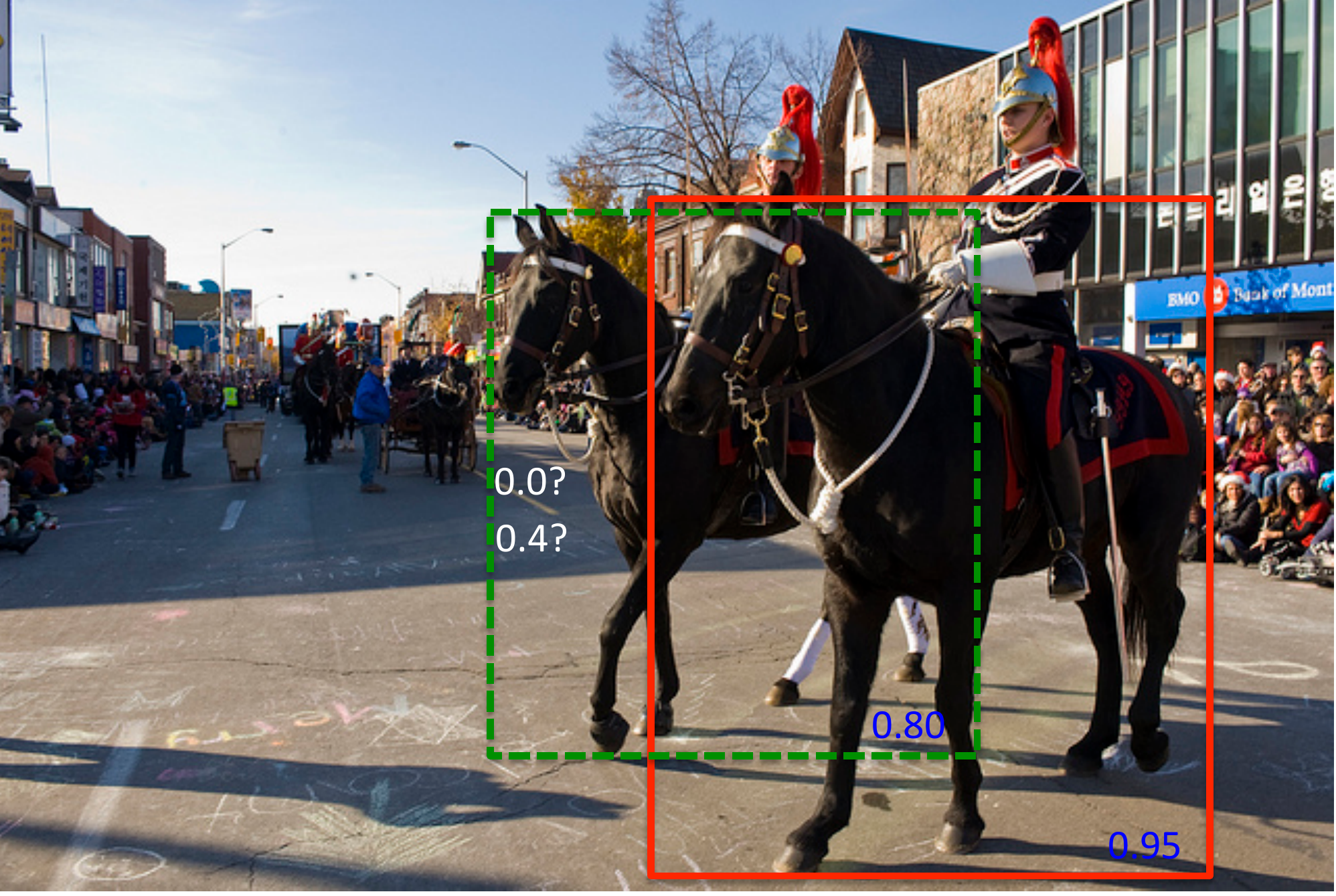}
\caption{This image has two confident horse detections (shown in red and green) which have a score of 0.95 and 0.8 respectively. The green detection box has a significant overlap with the red one. Is it better to suppress the green box altogether and assign it a score of 0 or a slightly lower score of 0.4?}
\label{fig:intro}
\end{figure}

%  \IncMargin{0.66em}
%  \begin{algorithm}
%  \SetKwData{Left}{left}\SetKwData{This}{this}\SetKwData{Up}{up}
%  \SetKwFunction{Union}{Union}\SetKwFunction{FindCompress}{FindCompress}
%  \SetKwInOut{Input}{Input}\SetKwInOut{Output}{Output}
%  \Input{$\mathcal{B} = \{b_1, ..,b_N\} $, $\mathcal{S} = \{s_1, ..,s_N\}$, $N_t$ \;
%  ~~~~~~~~~~~~~~~$\mathcal{B}$ is the list of initial detection boxes\; 
%  ~~~~~~~~~~~~~~~$\mathcal{S}$ contains corresponding detection scores\;
%  ~~~~~~~~~~~~~~~$N_t$ is the NMS threshold
%  }
%  \Begin{
%  $\mathcal{D} \leftarrow \{\}$ \;
%  \While{$\mathcal{B} \ne$ empty}{
%  $m \leftarrow$ argmax $\mathcal{S} $\;
%  $\mathcal{M} \leftarrow$ $b_m$\;
%  $\mathcal{D} \leftarrow \mathcal{D} \bigcup \mathcal{M }$; $ \mathcal{B} \leftarrow \mathcal{B} -  \mathcal{M}$\;
%  \For{$b_i ~ in ~ \mathcal{B}$} { 
%  \;
%   \textcolor{red}{
%   \If{~$iou(\mathcal{M}, b_i) \ge N_t$~}{
%   $ \mathcal{B} \leftarrow \mathcal{B} -  b_i$;~$ \mathcal{S} \leftarrow \mathcal{S} -  s_i$\;
%   }} 
%  \;
%   \textbf{\textcolor{mygreen}{$ s_i \leftarrow s_i f(iou(\mathcal{M}, b_i)) $\;}}  
%  \;
%   }
%  }
%  \Return{$\mathcal{D}, \mathcal{S}$}
%  }
%  \caption{The pseudo code in red is replaced with the one in green in Soft-NMS. We propose to revise the detection scores by scaling them as a linear or Gaussian function of overlap.}
%  \label{alg:nms}
%  \end{algorithm}\DecMargin{0em}

\begin{figure}
\centering
  \includegraphics[width=1\linewidth]{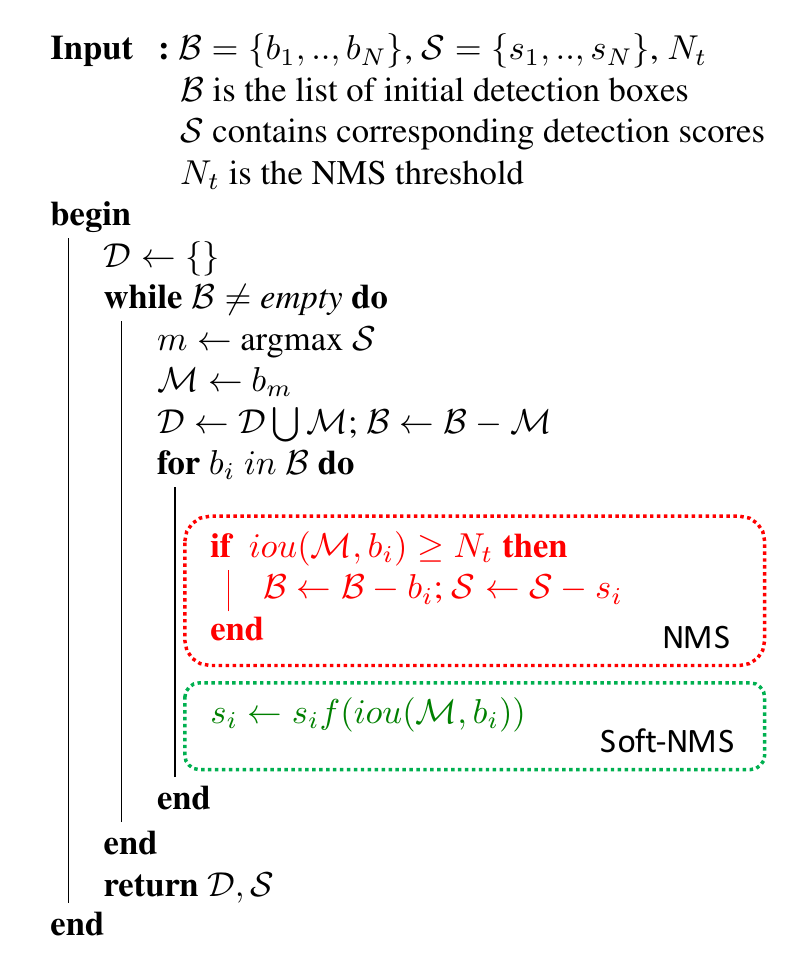}
\caption{The pseudo code in red is replaced with the one in green in Soft-NMS. We propose to revise the detection scores by scaling them as a linear or Gaussian function of overlap.}
\label{alg:nms}
\end{figure}

Traditional object detection pipelines \cite{dalal2005histograms, felzenszwalb2010object} employ a multi-scale sliding window based approach which assigns foreground/background scores for each class on the basis of features computed in each window. However, neighboring windows often have correlated scores (which increases false positives), so non-maximum suppression is used as a post-processing step to obtain final detections. With the advent of deep learning, the sliding window approach was replaced with category independent region proposals generated using a convolutional neural network. In state-of-the-art detectors, these proposals are input to a classification sub-network which assigns them class specific scores \cite{li2016r, ren2015faster}. Another parallel regression sub-network refines the position of these proposals. This refinement process improves localization for objects, but also leads to cluttered detections as multiple proposals often get regressed to the same region of interest (RoI). Hence, even in state-of-the-art detectors, non-maximum suppression is used to obtain the final set of detections as it significantly reduces the number of false positives.

Non-maximum suppression starts with a list of detection boxes $\mathcal{B}$ with scores $\mathcal{S}$. After selecting the detection with the maximum score $\mathcal{M}$, it removes it from the set $\mathcal{B}$ and appends it to the set of final detections $\mathcal{D}$. It also removes any box which has an overlap greater than a threshold $N_t$ with $\mathcal{M}$ in the set $\mathcal{B}$. This process is repeated for remaining boxes $\mathcal{B}$. A major issue with non-maximum suppression is that it sets the score for neighboring detections to zero. Thus, if an object was actually present in that overlap threshold, it would be missed and this would lead to a drop in average precision. However, if we lower the detection scores as a function of its overlap with $\mathcal{M}$, it would still be in the ranked list, although with a lower confidence. We show an illustration of the problem in Fig \ref{fig:intro}.

%whats missing in NMS?
Using this intuition, we propose a single line modification to the traditional greedy NMS algorithm in which we decrease the detection scores as an increasing function of overlap instead of setting the score to zero as in NMS. Intuitively, if a bounding box has a very high overlap with $\mathcal{M}$, it should be assigned a very low score, while if it has a low overlap, it can maintain its original detection score. This Soft-NMS algorithm is shown in Figure \ref{alg:nms}. Soft-NMS leads to noticeable improvements in average precision measured over multiple overlap thresholds for state-of-the-object detectors on standard datasets like PASCAL VOC and MS-COCO. Since Soft-NMS does not require any extra-training and is simple to implement, it can be easily integrated in the object detection pipeline.

\section{Related Work}
NMS has been an integral part of many detection algorithms in computer vision for almost 50 years. It was first employed in edge detection techniques \cite{rosenfeld1971edge}. Subsequently, it has been applied to multiple tasks like feature point detection \cite{lowe2004distinctive,harris1988combined,mikolajczyk2004scale}, face detection \cite{viola2001rapid} and object detection \cite{dalal2005histograms, felzenszwalb2010object,girshick2014rich}. In edge detection, NMS performs edge thinning to remove spurious responses \cite{rosenfeld1971edge,canny1986computational,zitnick2014edge}. In feature point detectors \cite{harris1988combined}, NMS is effective in performing local thresholding to obtain unique feature point detections. In face detection \cite{viola2001rapid}, NMS is performed by partitioning bounding-boxes into disjoint subsets using an overlap criterion. The final detections are obtained by averaging the co-ordinates of the detection boxes in the set. For human detection, Dalal and Triggs \cite{dalal2005histograms} demonstrated that a greedy NMS algorithm, where a bounding box with the maximum detection score is selected and its neighboring boxes are suppressed using a pre-defined overlap threshold improves performance over the approach used for face detection \cite{viola2001rapid}. Since then, greedy NMS has been the {\em de-facto} algorithm used in object detection \cite{felzenszwalb2010object,girshick2014rich,ren2015faster,li2016r}.

It is surprising that this component of the detection pipeline has remained untouched for more than a decade. Greedy NMS still obtains the best performance when average precision (AP) is used as an evaluation metric and is therefore employed in state-of-the-art detectors \cite{ren2015faster,li2016r}. A few learning-based methods have been proposed as an alternative to greedy NMS which obtain good performance for {\em object class detection} \cite{desai2011discriminative,rothe2014non,mrowca2015spatial}. For example, \cite{rothe2014non} first computes overlap between each pair of detection boxes. It then performs affinity propagation clustering to select exemplars for each cluster which represent the final detection boxes. A multi-class version of this algorithm is proposed in \cite{mrowca2015spatial}. However, object class detection is a different problem, where object instances of all classes are evaluated simultaneously per image. Hence, we need to select a threshold for all classes and generate a fixed set of boxes. Since different thresholds may be suitable for different applications, in generic object detection, average precision is computed using a ranked list of all object instances in a particular class. Therefore, greedy NMS performs favourably to these algorithms on generic object detection metrics. 

In another line of work, for detecting salient objects, a proposal subset optimization algorithm was proposed \cite{zhang2016unconstrained} as an alternative to greedy NMS. It performs a MAP-based subset optimization to jointly optimize the number and locations of detection
windows. In salient object detection, the algorithm is expected to only find salient objects and not all objects. So, this problem is also different from generic object detection and again greedy NMS performs favourably when performance on object detection metrics is measured. For special cases like pedestrian detection, a quadratic unconstrained binary optimization (QUBO) solution was proposed which uses detection scores as a unary potential and overlap between detections as a pairwise potential to obtain the optimal subset of detection boxes \cite{rujikietgumjorn2013optimized}. Like greedy NMS, QUBO also applies a hard threshold to suppress detection boxes, which is different from Soft-NMS. In another learning-based framework for pedestrian detection, a determinantal point process was combined with individualness prediction scores to optimally select final detections \cite{lee2016individualness}. To the best of our knowledge, for generic object detection, greedy NMS is still the strongest baseline on challenging object detection datasets like PASCAL VOC and MS-COCO.

\begin{figure}
\centering
   \includegraphics[width=1\linewidth]{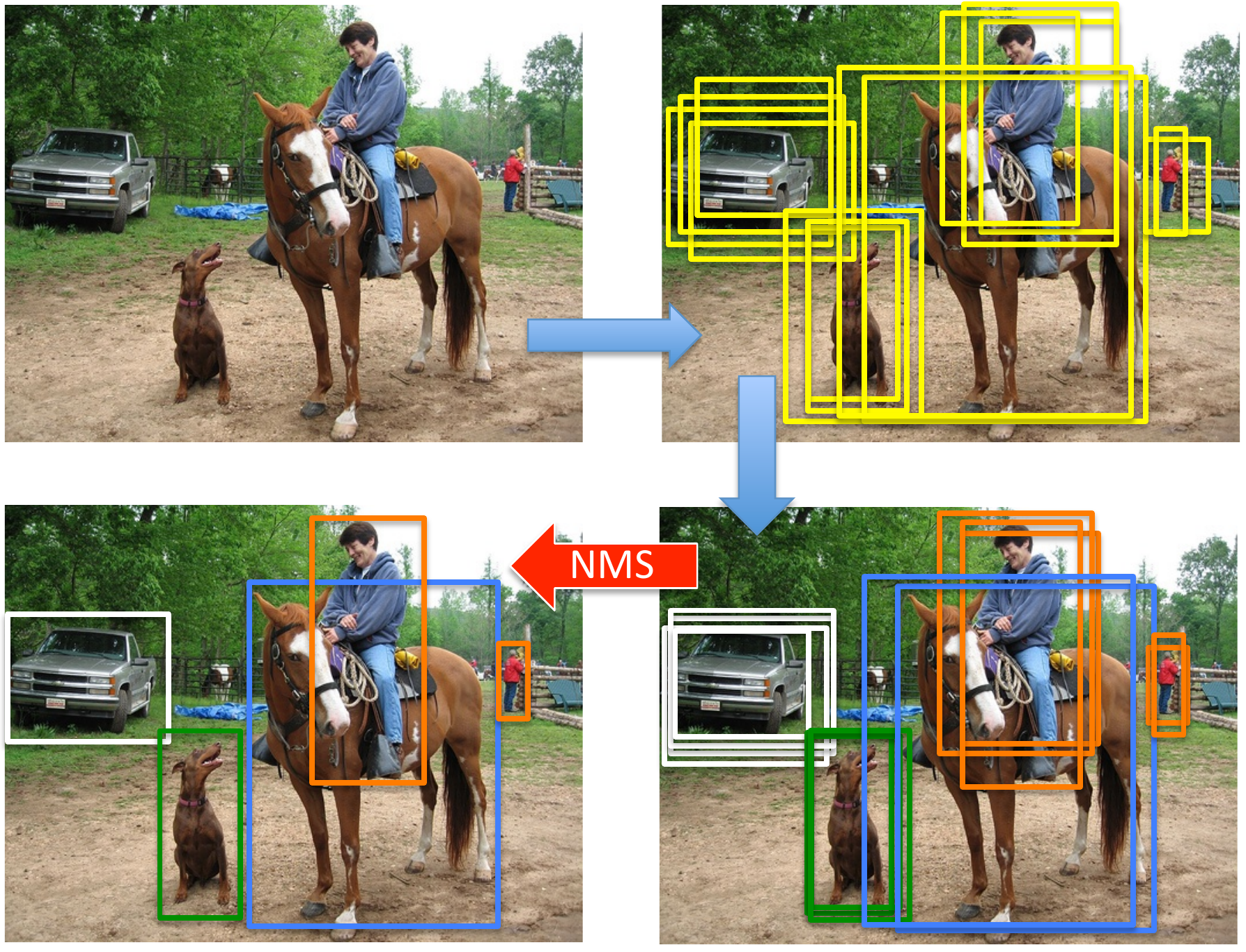}
\caption{In object detection, first category independent region proposals are generated. These region proposals are then assigned a score for each class label using a classification network and their positions are updated slightly using a regression network. Finally, non-maximum-suppression is applied to obtain detections.}
\label{fig:bg}
\end{figure}

\section{Background}
We briefly describe the object-detection pipeline used in state-of-the-art object detectors in this section. During inference, an object detection network performs a sequence of convolution operations on an image using a deep convolutional neural network (CNN). The network bifurcates into two branches at a layer $L$ --- one branch generates region proposals while the other performs classification and regression by pooling convolutional features inside RoIs generated by the proposal network. The proposal network generates classification scores and regression offsets for anchor boxes of multiple scales and aspect ratios placed at each pixel in the convolutional feature map \cite{ren2015faster}. It then ranks these anchor boxes and selects the top $K$ ($\approx$ 6000) anchors to which the bounding box regression offsets are added to obtain image level co-ordinates for each anchor. Greedy non-maximum suppression is applied to top $K$ anchors which eventually generates region proposals \footnote{We do not replace this non-maximum suppression in the object detection pipeline.}. 

The classification network generates classification and regression scores for each proposal generated by the proposal network. Since there is no constraint in the network which forces it to generate a unique RoI for an object, multiple proposals may correspond to the same object. Hence, other than the first correct bounding-box, all other boxes on the same object would generate false positives. To alleviate this problem, non-maximum-suppression is performed on detection boxes of each class independently, with a specified overlap threshold. Since the number of detections is typically small and can be further reduced by pruning detections which fall below a very small threshold, applying non-maximum suppression at this stage is not computationally expensive. We present an alternative approach to this non-maximum suppression algorithm in the object detection pipeline. An overview of the object detection pipeline is shown in Fig \ref{fig:bg}.

\section{Soft-NMS} \label{sec:snms}
Current detection evaluation criteria emphasise precise localization and measure average precision of detection boxes at multiple overlap thresholds (ranging from 0.5 to 0.95). Therefore, applying NMS with a low threshold like 0.3 could lead to a drop in average precision when the overlap criterion during evaluation for a true positive is 0.7 (we refer to the detection evaluation threshold as $O_t$ from here on). This is because, there could be a detection box $b_i$ which is very close to an object (within 0.7 overlap), but had a slightly lower score than $\mathcal{M}$ ($\mathcal{M}$ did not cover the object), thus $b_i$ gets suppressed by a low $N_t$. The likelihood of such a case would increase as the overlap threshold criterion is increased. Therefore, suppressing all nearby detection boxes with a low $N_t$ would increase the miss-rate. 

Also, using a high $N_t$ like 0.7 would increase false positives when $O_t$ is lower and would hence drop precision averaged over multiple thresholds. The increase in false positives would be much higher than the increase in true positives for this case because the number of objects is typically much smaller than the number of RoIs generated by a detector. Therefore, using a high NMS threshold is also not optimal. 

To overcome these difficulties, we revisit the NMS algorithm in greater detail. The pruning step in the NMS algorithm can be written as a re-scoring function as follows, 

\[
    s_i= 
\begin{cases}
    s_i,& \textrm{iou}(\mathcal{M}, b_i) < N_t\\
    0,              &  \textrm{iou}(\mathcal{M}, b_i) \ge N_t
\end{cases}
\;,
\]

Hence, NMS sets a hard threshold while deciding what should be kept or removed from the neighborhood of $\mathcal{M}$. Suppose, instead, we decay the classification score of a box $b_i$ which has a high overlap with $\mathcal{M}$, rather than suppressing it altogether. If $b_i$ contains an object not covered by $\mathcal{M}$, it won't lead to a miss at a lower detection threshold. However, if $b_i$ does not cover any other object (while $\mathcal{M}$ covers an object), and even after decaying its score it ranks above true detections, it would still generate a false positive. Therefore, NMS should take the following conditions into account,

\begin{itemize}
    \item Score of neighboring detections should be decreased to an extent that they have a smaller likelihood of increasing the false positive rate, while being above obvious false positives in the ranked list of detections. 
    \item Removing neighboring detections altogether with a low NMS threshold would be sub-optimal and would increase the miss-rate when evaluation is performed at high overlap thresholds.
    \item Average precision measured over a range of overlap thresholds would drop when a high NMS threshold is used.
\end{itemize}

We evaluate these conditions through experiments in Section \ref{sec:analysis}. 

\textbf{Rescoring Functions for Soft-NMS:} Decaying the scores of other detection boxes which have an overlap with $\mathcal{M}$ seems to be a promising approach for improving NMS. It is also clear that scores for detection boxes which have a higher overlap with $\mathcal{M}$ should be decayed more, as they have a higher likelihood of being false positives. Hence, we propose to update the pruning step with the following rule,

\[
    s_i= 
\begin{cases}
    s_i,& \textrm{iou}(\mathcal{M}, b_i) < N_t\\
    s_i (1 - \textrm{iou}(\mathcal{M}, b_i)),              &  \textrm{iou}(\mathcal{M}, b_i) \ge N_t
\end{cases}
\;,
\]
The above function would decay the scores of detections above a threshold $N_t$ as a linear function of overlap with $\mathcal{M}$. Hence, detection boxes which are far away from $\mathcal{M}$ would not be affected and those which are very close would be assigned a greater penalty. 

However, it is not continuous in terms of overlap and a sudden penalty is applied when a NMS threshold of $N_t$ is reached. It would be ideal if the penalty function was continuous, otherwise it could lead to abrupt changes to the ranked list of detections. A continuous penalty function should have no penalty when there is no overlap and very high penalty at a high overlap. Also, when the overlap is low, it should increase the penalty gradually, as $\mathcal{M}$ should not affect the scores of boxes which have a very low overlap with it. However, when overlap of a box $b_i$ with $\mathcal{M}$ becomes close to one, $b_i$ should be significantly penalized. Taking this into consideration, we propose to update the pruning step with a Gaussian penalty function as follows,

$$s_i = s_i e^{-\frac{\textrm{iou}(\mathcal{M}, b_i)^2}{\sigma}},  \forall b_i \notin \mathcal{D}$$
This update rule is applied in each iteration and scores of {\em all remaining} detection boxes are updated. 

The Soft-NMS algorithm is formally described in Figure \ref{alg:nms}, where $f(iou(\mathcal{M},b_i)))$ is the overlap based weighting function. The computational complexity of each step in Soft-NMS is $\mathcal{O}(N)$, where $N$ is the number of detection boxes. This is because scores for all detection boxes which have an overlap with $\mathcal{M}$ are updated. So, for $N$ detection boxes, the computational complexity for Soft-NMS is $\mathcal{O}(N^2)$, which is the same as traditional greedy-NMS. Since NMS is not applied on all detection boxes (boxes with a minimum threshold are pruned in each iteration), this step is not computationally expensive and hence does not affect the running time of current detectors.

\begin{table*}[t]
\begin{center}
%\small
\resizebox{\textwidth}{!}{
\begin{tabular}{|c|c|c|c|c|c|c|c|c|c|}
  \hline
  
  Method & Training data & Testing data & AP 0.5:0.95 & AP @ 0.5 & AP small & AP medium & AP large & Recall @ 10 & Recall @ 100\\
  \hline\hline
  
  R-FCN \cite{li2016r}  & train+val35k & test-dev & 31.1& 52.5  & 14.4 & 34.9 & 43.0 & 42.1 & 43.6\\
  R-FCN + S-NMS G  & train+val35k & test-dev & \textbf{32.4} & \textbf{53.4}  & \textbf{15.2} & \textbf{36.1} & \textbf{44.3} & \textbf{46.9} & \textbf{52.0}\\
  R-FCN + S-NMS L  & train+val35k & test-dev & 32.2 & \textbf{53.4} & 15.1 & 36.0 & 44.1 & 46.0 & 51.0\\
  \hline
  
  F-RCNN \cite{ren2015faster}  & train+val35k & test-dev & 24.4 & 45.7  & 7.9 & 26.6 & 37.2 & 36.1 & 37.1\\
  F-RCNN + S-NMS G & train+val35k & test-dev & \textbf{25.5} & 46.6  & \textbf{8.8} & \textbf{27.9} & \textbf{38.5} & \textbf{41.2} & 45.3\\
    
  F-RCNN + S-NMS L & train+val35k & test-dev & \textbf{25.5} &\textbf{46.7}  & \textbf{8.8} & \textbf{27.9} & 38.3 & 40.9 & \textbf{45.5}\\
  \hline
 
  D-RFCN \cite{dai2017deformable} & trainval & test-dev & 37.4 & 59.6 & 17.8 & 40.6 & 51.4 & 46.9 & 48.3 \\
  D-RFCN S-NMS G & trainval & test-dev & \textbf{38.4} & \textbf{60.1} & \textbf{18.5} & \textbf{41.6} & \textbf{52.5} & \textbf{50.5} & \textbf{53.8} \\
  D-RFCN + MST & trainval & test-dev & 39.8 & 62.4 & 22.6 & 42.3 & 52.2 & 50.5 & 52.9 \\
  D-RFCN + MST + S-NMS G & trainval & test-dev & \textbf{40.9} & \textbf{62.8} & \textbf{23.3} & \textbf{43.6} & \textbf{53.3} & \textbf{54.7} & \textbf{60.4}\\
  \hline

 \end{tabular}
 }
\caption{Results on MS-COCO test-dev set for R-FCN, D-RFCN and Faster-RCNN (F-RCNN) which use NMS as baseline and our proposed Soft-NMS method. G denotes Gaussian weighting and L denotes linear weighting. MST denotes multi-scale testing.}
\label{tab:coco}
\end{center}
\end{table*}

 \begin{table*}[t]
 \begin{center}
 %\small
 \resizebox{\textwidth}{!}{
 \begin{tabular}{|c|c|c|cccccccccccccccccccc|}
   \hline
   \ct{Method} & \ct{AP} & \ct{AP@0.5} & \ct{areo} & \ct{bike} & \ct{bird} & \ct{boat} & \ct{bottle} & \ct{bus} & \ct{car} & \ct{cat} & \ct{chair} & \ct{cow} & \ct{table} & \ct{dog} & \ct{horse} & \ct{mbike} & \ct{person} & \ct{plant} & \ct{sheep} & \ct{sofa} & \ct{train} & \ct{tv} \\
   \hline\hline
  
 \cite{ren2015faster}+NMS & 37.7 & 70.0 & 37.8 & 44.6 & 34.7 & 24.4 & 23.4 & 50.6 & 50.1 & 45.1 & 25.1 & 42.6 & 36.5 & 40.7 & 46.8 & 39.8 & 38.2 & 17.0 		                 		 & 37.8 & 36.4 & 43.7 & \textbf{38.9} \\
                             \cite{ren2015faster}+S-NMS G & \textbf{39.4} & \textbf{71.2} & 40.2 & \textbf{46.6} & \textbf{36.7} & 25.9 & \textbf{24.9} & \textbf{51.9} & 51.6 & \textbf{48.0} & \textbf{25.3} & 44.5 & \textbf{37.3} & 42.6 & \textbf{49.0} & 42.2 & 41.6 & 17.7 & 39.2 & \textbf{39.3} & \textbf{45.9} & 37.5 \\
                              
                            \cite{ren2015faster}+S-NMS L & \textbf{39.4} &
                            \textbf{71.2} &
                            \textbf{40.3} & \textbf{46.6} & 36.3 & \textbf{27.0} & 24.2 & 51.2 & \textbf{52.0} & 47.2 & \textbf{25.3} & \textbf{44.6} & 37.2 & \textbf{45.1} & 48.3 & \textbf{42.3} & \textbf{42.3} & \textbf{18.0} & \textbf{39.4} & 37.1 & 45.0 & 38.7 \\
\cite{li2016r}+NMS & 49.8 & 79.4 & 52.8 & 54.4 & 47.1 & 37.6 & 38.1 & 63.4 & 59.4 & 62.0 & 35.3 & 56.0 & 38.9 & 59.0 & 54.5 & 50.5 & 47.6 & 24.8 & 53.3 & 52.2 
 					   & 57.4 & 52.7 \\
                       \cite{li2016r}+S-NMS G & 51.4 & \textbf{80.0} &  \textbf{53.8} & \textbf{56.0} & 48.3 & 39.9 & 39.4 & \textbf{64.7} & 61.3 & 64.7 & \textbf{36.3} & \textbf{57.0} & \textbf{40.2} & 60.6 & 55.5 & 52.1 & \textbf{50.7} & \textbf{26.5} & 53.8 & 53.5 & 59.3 & 53.8 \\
                       
                       \cite{li2016r}+S-NMS L & \textbf{51.5} & \textbf{80.0} & 53.2 & 55.8 & \textbf{48.9} & \textbf{40.0} & \textbf{39.6} & 64.6 & \textbf{61.5} & \textbf{65.0} & \textbf{36.3} & 56.5 & \textbf{40.2} & \textbf{61.3} & \textbf{55.6} & \textbf{52.9} & 50.3 & 26.2 & \textbf{54.3} & \textbf{53.6} & \textbf{59.5} & \textbf{53.9} \\ 
 \hline

 \end{tabular}
 }
 \caption{Results on Pascal VOC 2007 test set for off-the-shelf standard object detectors which use NMS as baseline and our proposed Soft-NMS method. Note that COCO-style evaluation is used. }
 \label{tab:pascal}
 \end{center}
 \end{table*}
 
Note that Soft-NMS is also a greedy algorithm and does not find the globally optimal re-scoring of detection boxes. Re-scoring of detection boxes is performed in a greedy fashion and hence those detections which have a high local score are not suppressed. However, Soft-NMS is a {\em generalized} version of non-maximum suppression and traditional NMS is a special case of it with a discontinuous binary weighting function. Apart from the two proposed functions, other functions with more parameters can also be explored with Soft-NMS which take overlap and detection scores into account. For example, instances of the generalized logistic function like the Gompertz function can be used, but such functions would increase the number of hyper-parameters.

\section{Datasets and Evaluation}

We perform experiments on two datasets, PASCAL VOC \cite{everingham2010pascal} and MS-COCO \cite{lin2014microsoft}. The Pascal dataset has 20 object categories, while the MS-COCO dataset has 80 object categories. We choose the VOC 2007 test partition to measure performance. For the MS-COCO dataset, sensitivity analysis is conducted on a publicly available minival set of 5,000 images. We also show results on the test-dev partition on the MS-COCO dataset which consists of 20,288 images.
\begin{table*}[t]
\begin{center}
 \begin{tabular}{|c|c|c|c|c|c|c|c|c|c|}
  \hline
  $N_t$ & AP @ 0.5 & AP @ 0.6 & AP @ 0.7 & AP @ 0.8 & $\sigma$ & AP @ 0.5 & AP @ 0.6 & AP @ 0.7 & AP @ 0.8 \\
  \hline
0.3 & 0.5193 & 0.4629 &  0.3823 &  0.2521 & 0.1 & 0.5202  &  0.4655  &  0.3846  &  0.2533  \\
\hline
0.4 & \textbf{0.5238} & 0.4680 &  0.3840 &  0.2524 & 0.3 & \textbf{0.5293}  &  0.4765  &  0.3960  &  0.2619  \\ 
\hline
0.5 & 0.5227 & \textbf{0.4708} &  0.3869 &  0.2526 & 0.5 & 0.5274  &  \textbf{0.4777}  &  0.3997  &  0.2669  \\ 
\hline
0.6 & 0.5127 & 0.4690 &  \textbf{0.3895} &  0.2527 & 0.7 & 0.5232  &  0.4757  &  \textbf{0.4001}  &  0.2695  \\
\hline
0.7 & 0.4894 & 0.4535 &  0.3860 &  \textbf{0.2535} & 0.9 & 0.5186  &  0.4727  &  0.3992  &  0.2710  \\
\hline
0.8 & 0.4323 & 0.4048 &  0.3569 &  0.2520 & 1.1 & 0.5136  &  0.4691  &  0.3976  &  \textbf{0.2713}  \\
  \hline 
 \end{tabular}
 
\caption{Sensitivity Analysis across multiple overlap thresholds $N_t$ and parameters $\sigma$ for NMS and Soft-NMS using R-FCN on coco minival. Best performance at each $O_t$ is marked in bold for each method.}
\label{tab:sens-rfcn}
\end{center}
\end{table*}

To evaluate our method, we experimented with three state-of-the-art detectors, namely, Faster-RCNN \cite{ren2015faster}, R-FCN \cite{li2016r} and Deformable-RFCN. For the PASCAL dataset, we selected publicly available pre-trained models provided by the authors. The Faster-RCNN detector was trained on VOC 2007 train set while the R-FCN detector was trained on VOC 2007 and 2012. For MS-COCO also, we use the publicly available model for Faster-RCNN. However, since there was no publicly available model trained on MS-COCO for R-FCN, we trained our own model in Caffe \cite{jia2014caffe} starting from a ResNet-101 CNN architecture \cite{he2016deep}. Simple modifications like 5 scales for RPN anchors, a minimum image size of 800, 16 images per minibatch and 256 ROIs per image were used. Training was done on 8 GPUs in parallel. Note that our implementation obtains 1.9\% better accuracy than that reported in \cite{li2016r} without using multi-scale training or testing. Hence, this is a strong baseline for R-FCN on MS-COCO. Both these detectors use a default NMS threshold of 0.3. In the sensitivity analysis section, we also vary this parameter and show results. We also trained deformable R-FCN with the same settings. At a threshold of 10e-4, using 4 CPU threads, it takes 0.01s per image for 80 classes. After each iteration, detections which fall below the threshold are discarded. This reduces computation time. At 10e-2, run time is 0.005 seconds on a single core. We set maximum detections per image to 400 on MS-COCO and the evaluation server selects the top 100 detections per class for generating metrics (we confirmed that the coco evaluation server was not selecting top 100 scoring detections per image till June 2017). Setting maximum detections to 100 reduces coco-style AP by 0.1.

\section{Experiments}
In this section, we show comparative results and perform sensitivity analysis to show robustness of Soft-NMS compared to traditional NMS. We also conduct specific experiments to understand why and where does Soft-NMS perform better compared to traditional NMS.

\subsection{Results}
In Table \ref{tab:coco} we compare R-FCN and Faster-RCNN with traditional non-maximum suppression and Soft-NMS on MS-COCO. We set $N_t$ to 0.3 when using the linear weighting function and $\sigma$ to 0.5 with the Gaussian weighting function. It is clear that Soft-NMS (using both Gaussian and linear weighting function) improves performance in all cases, especially when AP is computed at multiple overlap thresholds and averaged. For example, we obtain an improvement of 1.3\% and 1.1\% respectively for R-FCN and Faster-RCNN, which is significant for the MS-COCO dataset. Note that we obtain this improvement by just changing the NMS algorithm and hence it can be applied easily on multiple detectors with minimal changes. We perform the same experiments on the PASCAL VOC 2007 test set, shown in Table  \ref{tab:coco}. We also report average precision averaged over multiple overlap thresholds like MS-COCO. Even on PASCAL VOC 2007, Soft-NMS obtains an improvement of 1.7\% for both Faster-RCNN and R-FCN. For detectors like SSD \cite{liu2016ssd} and YOLOv2 \cite{redmon2016you} which are not proposal based, with the linear function, Soft-NMS only obtains an improvement of 0.5\%. This is because proposal based detectors have higher recall and hence Soft-NMS has more potential to improve recall at higher $O_t$.

\begin{figure}
    \centering
    \includegraphics[width=0.49\linewidth]{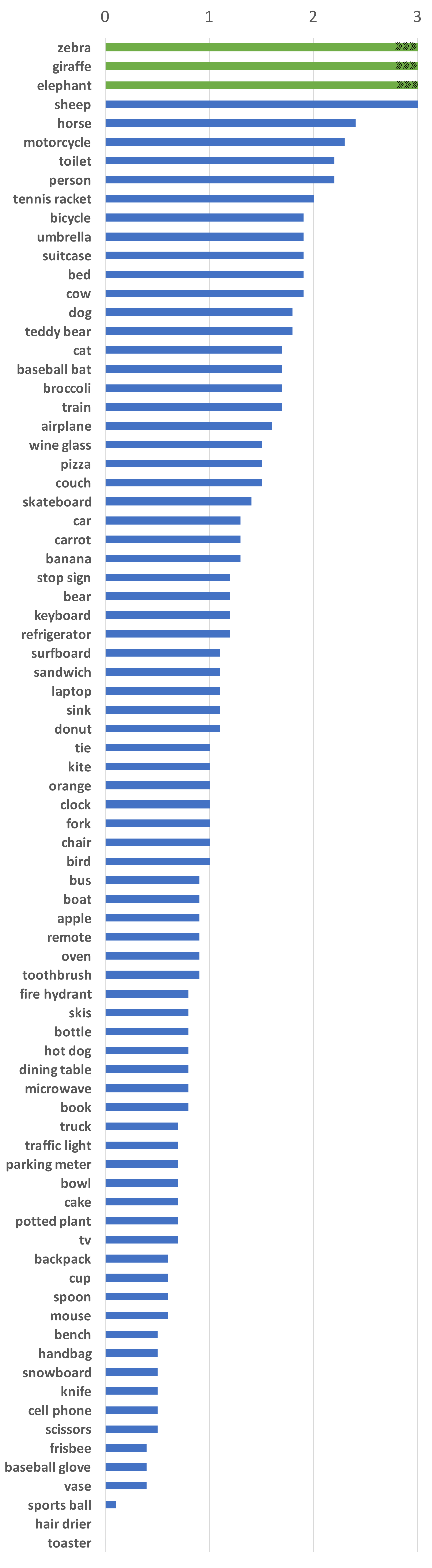} 
    \includegraphics[width=0.49\linewidth]{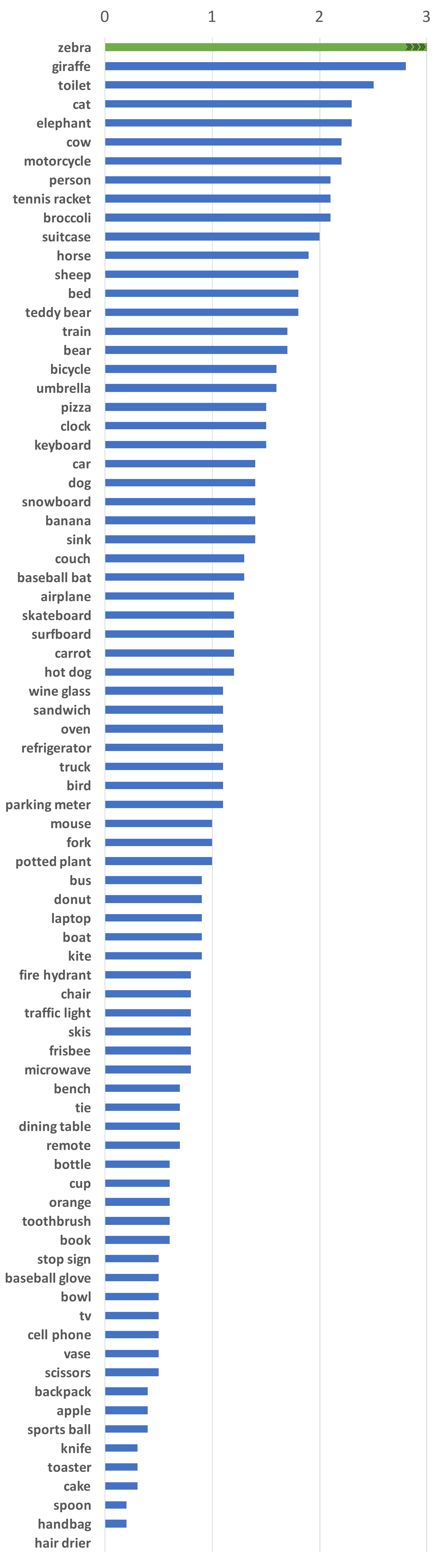}
    \caption{Per class improvement in AP for MS-COCO using Soft-NMS for R-FCN is shown in the left and for Faster-RCNN is shown on the right. Green bars indicate improvements beyond 3 \%}
    \label{fig:per-class}
    \vspace{-0.2in}
\end{figure}

From here on, in all experiments, when we refer to Soft-NMS, it uses the Gaussian weighting function. In Fig \ref{fig:per-class}, we also show per-class improvement on MS-COCO. It is interesting to observe that Soft-NMS when applied on R-FCN improves maximum performance for animals which are found in a herd like zebra, giraffe, sheep, elephant, horse by 3-6\%, while there is little gain for objects like toaster, sports ball, hair drier which are less likely to co-occur in the same image. 
\begin{figure}
    \centering
    \includegraphics[width=1\linewidth]{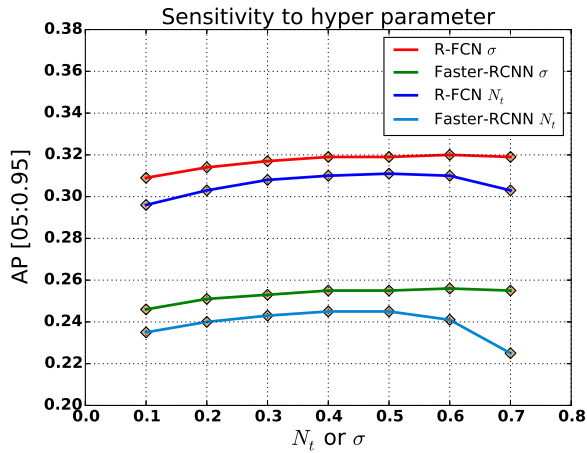} 
    \caption{R-FCN Sensitivity to hyper parameters $\sigma$ (Soft-NMS) and $N_t$ (NMS) }
    \label{fig:sen-rfcn-sfnms}
\end{figure}

%\begin{figure*}[tb]
%    \centering
%    \includegraphics[width=1\linewidth]{pics/qual_pics/failed.pdf} 
%    \caption{Qualitative results failure cases: Image pairs are shown %in which the left image with blue bounding boxes is for traditional %NMS, while the right image with red bounding boxes is for Soft-NMS.}
%    \label{fig:qualitative-fail}
%\end{figure*}
\subsection{Sensitivity Analysis}
Soft-NMS has a $\sigma$ parameter and traditional NMS has an overlap threshold parameter $N_t$. We vary these parameters and measure average precision on the minival set of MS-COCO set for each detector, see Fig \ref{fig:sen-rfcn-sfnms}. Note that AP is stable between 0.3 to 0.6 and drops significantly outside this range for both detectors. The variation in AP in this range is around 0.25\% for traditional NMS. Soft-NMS obtains better performance than NMS from a range between 0.1 to 0.7. Its performance is stable from 0.4 to 0.7 and better by $\sim$1\% for each detector even on the best NMS threshold selected by us on the coco-minival set. In all our experiments, we set $\sigma$ to 0.5, even though a $\sigma$ value of 0.6 seems to give better performance on the coco minival set. This is because we conducted the sensitivity analysis experiments later on and a difference of 0.1\% was not significant.
\begin{figure*}[tb]
    \centering
    \includegraphics[width=0.33\linewidth]{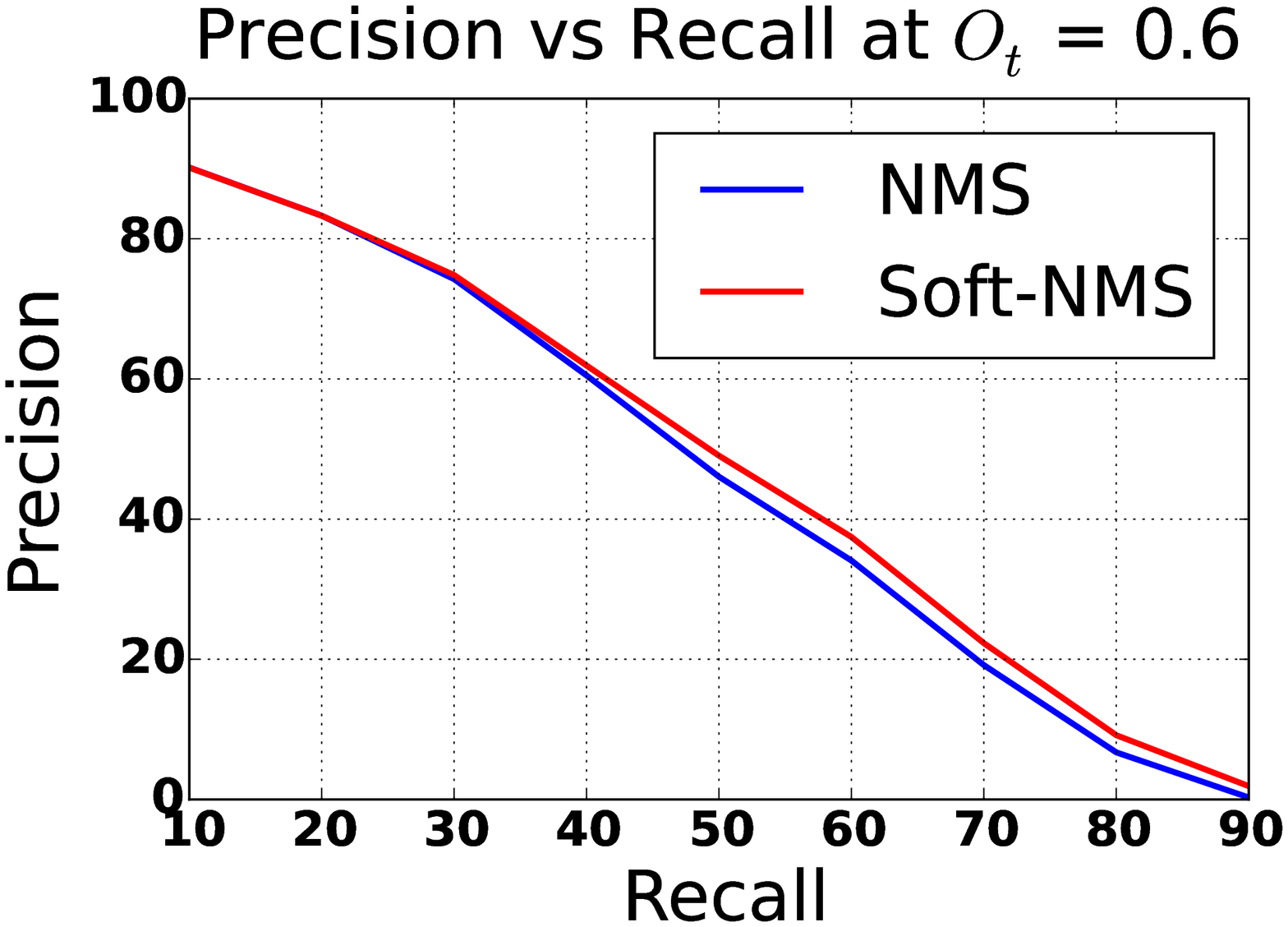} 
    \includegraphics[width=0.33\linewidth]{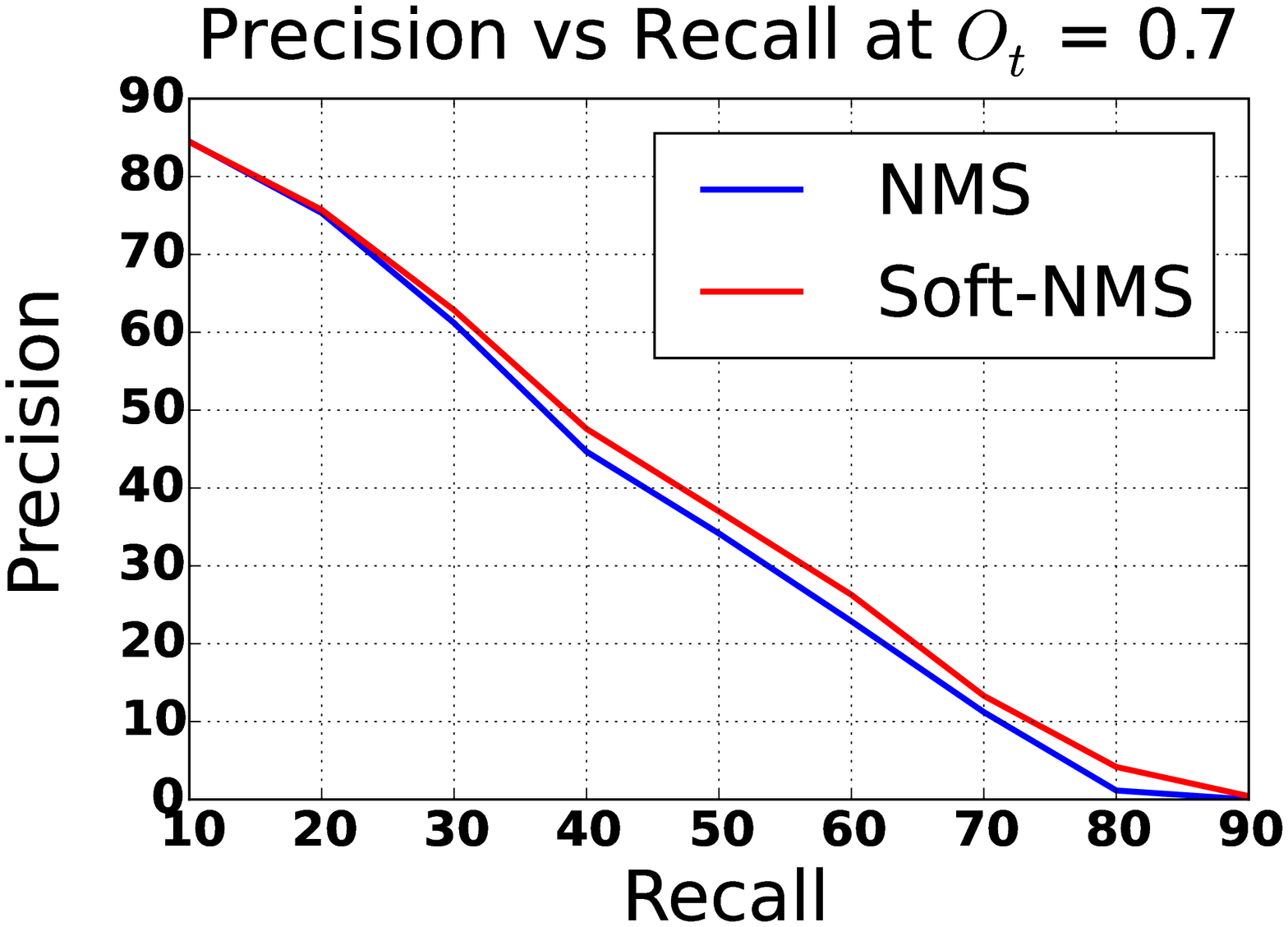} 
    \includegraphics[width=0.33\linewidth]{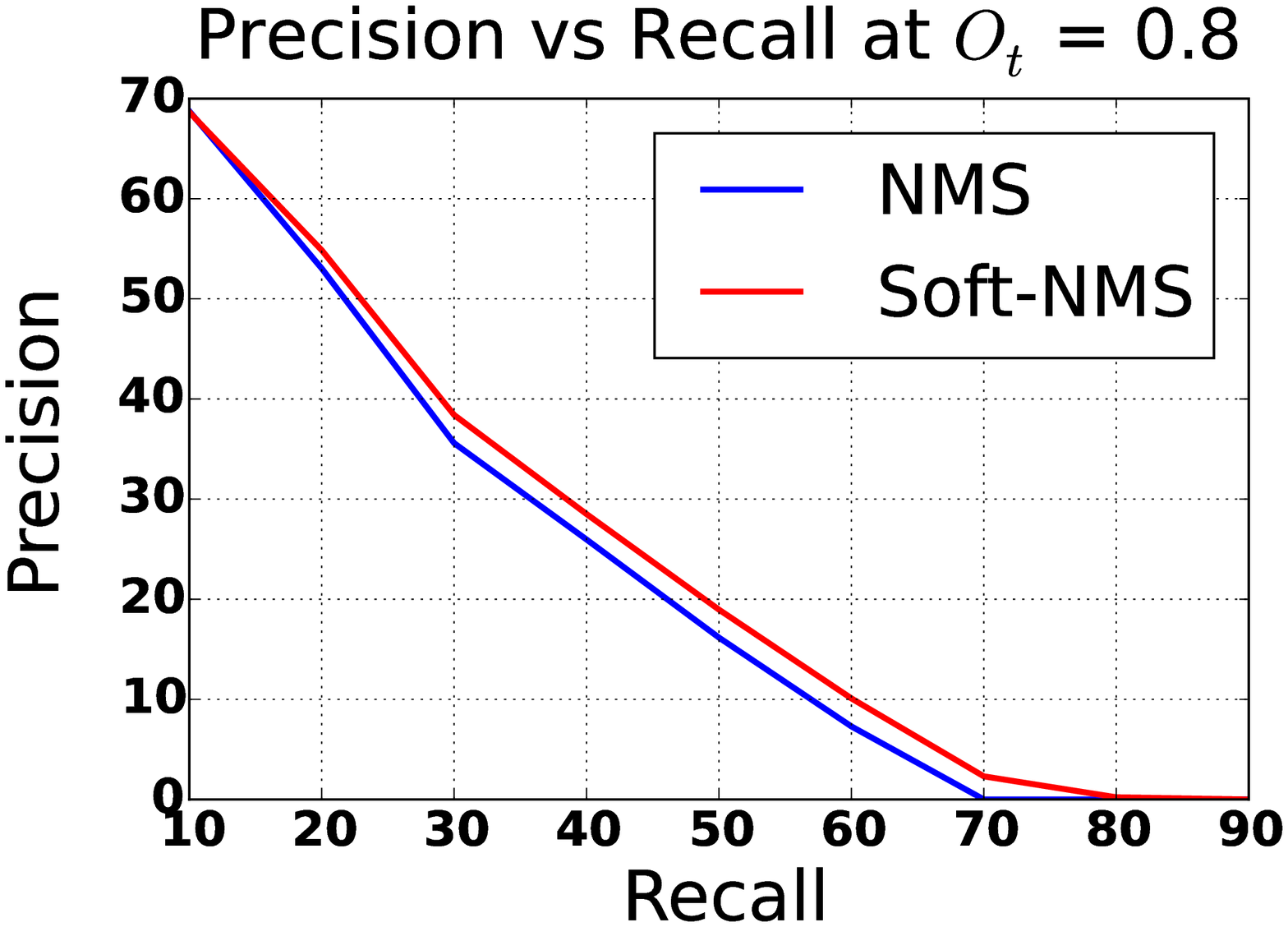} 
    \caption{R-FCN : Precision vs Recall at multiple overlap thresholds $O_t$}
    \label{fig:prfc-def}
\end{figure*}
\subsection{When does Soft-NMS work better?} \label{sec:analysis}
\textbf{Localization Performance}
Average precision alone does not explain us clearly when Soft-NMS obtains significant gains in performance. Hence, we present average precision of NMS and Soft-NMS when measured at different overlap thresholds. We also vary the NMS and Soft-NMS hyper-parameters to understand the characteristics of both these algorithms.
From Table \ref{tab:sens-rfcn}, we can infer that average precision decreases as NMS threshold is increased. Although it is the case that for a large $O_t$, a high $N_t$ obtains slightly better performance compared to a lower $N_t$ --- AP does not drop significantly when a lower $N_t$ is used. On the other hand, using a high $N_t$ leads to significant drop in AP at lower $O_t$ and hence when AP is averaged at multiple thresholds, we observe a performance drop. Therefore, a better performance using a higher $N_t$ does not generalize to lower values of $O_t$ for traditional NMS.

However, when we vary $\sigma$ for Soft-NMS, we observe a different characteristic. Table \ref{tab:sens-rfcn} shows that even when we obtain better performance at higher $O_t$, performance at lower $O_t$ does not drop. Further, we observe that Soft-NMS performs significantly better ($\sim$2\%) than traditional NMS irrespective of the value of the selected $N_t$ at higher $O_t$. Also, the best AP for any hyper-parameter ($N_t$ or $\sigma$) for a selected $O_t$ is always better for Soft-NMS. This comparison makes it very clear that across all parameter settings, the best $\sigma$ parameter for Soft-NMS performs better than a hard threshold $N_t$ selected in traditional NMS. Further, when performance across all thresholds is averaged, since a single parameter setting in Soft-NMS works well at multiple values of $O_t$, overall performance gain is amplified. As expected, low values of $\sigma$ perform better at lower $O_t$ and higher values of sigma perform better at higher $O_t$. Unlike NMS, where higher values of $N_t$ lead to very little improvement in AP, higher values of $\sigma$ lead to significant improvement in AP at a higher $O_t$.
Therefore, a larger $\sigma$ can be used to improve performance of the detector for better localization which is not the case with NMS, as a larger $N_t$ obtains very little improvement.

\textbf{Precision vs Recall}
Finally, we would like to also know at what recall values is Soft-NMS performing better than NMS at different $O_t$. Note that we re-score the detection scores and assign them lower scores, so we do not expect precision to improve at a lower recall. However, as $O_t$ and recall is increased, Soft-NMS obtains significant gains in precision. This is because, traditional NMS assigns a zero score to all boxes which have an overlap greater than $N_t$ with $\mathcal{M}$. Hence, many boxes are missed and therefore precision does not increase at higher values of recall. Soft-NMS re-scores neighboring boxes instead of suppressing them altogether which leads to improvement in precision at higher values of recall. Also, Soft-NMS obtains significant improvement even for lower values of recall at higher values of $O_t$ because near misses are more likely to happen in this setting.
\begin{figure*}[b]
    \centering
    \includegraphics[width=1\linewidth]{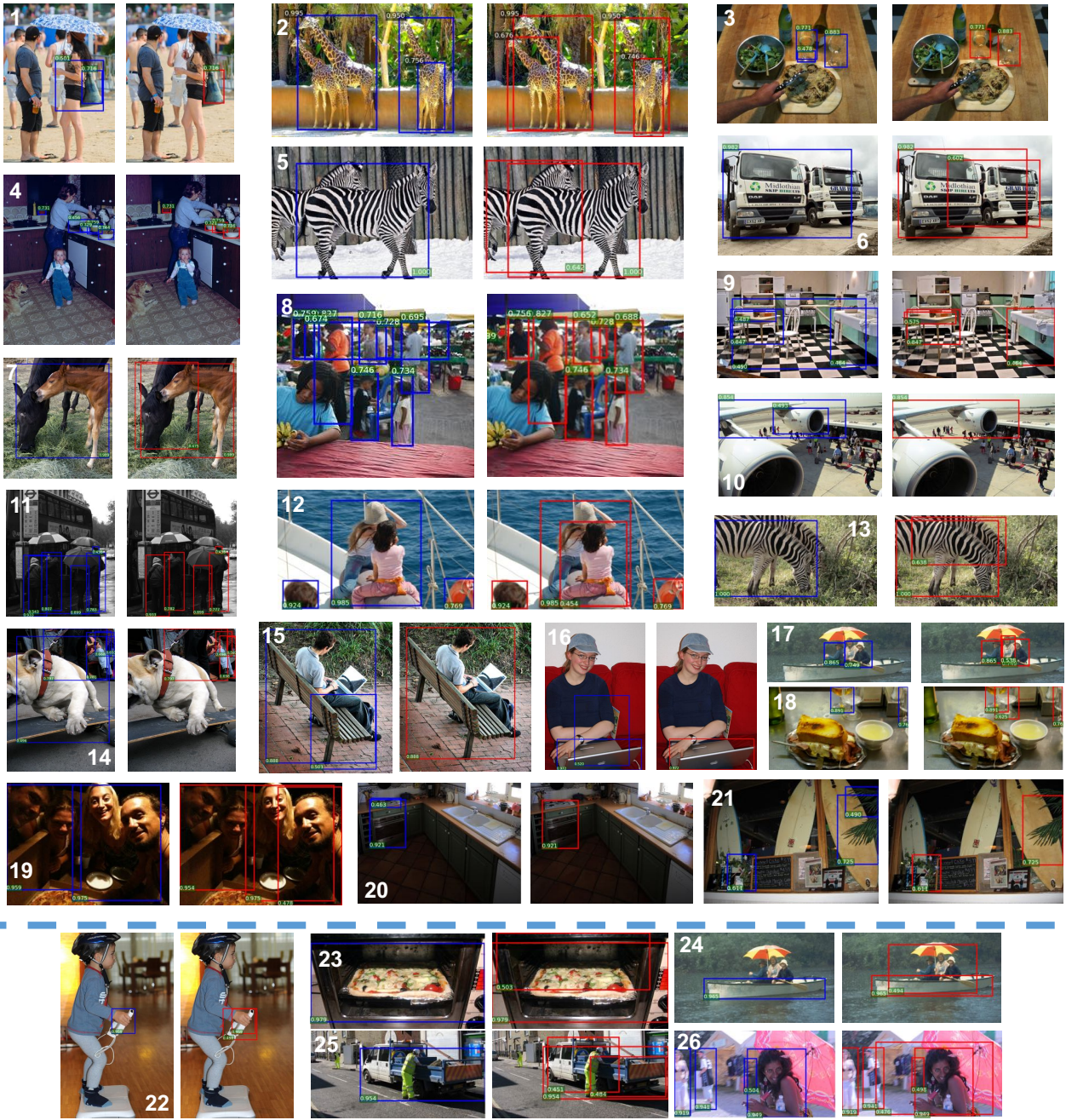} 
    \caption{Qualitative results: Image pairs are shown in which the left image with blue bounding boxes is for traditional NMS, while the right image with red bounding boxes is for Soft-NMS. Examples above the blue line are successful cases and below are failure cases. Even in failure cases, Soft-NMS assigns a significantly lower score than the bounding box with a higher score. Note that for image No. 14, the classes in consideration are for ``person" and not ``dog", similarly in 15 it is ``bench" and in 21 it is ``potted plant".}
    \label{fig:qualitative}
\end{figure*}
\subsection{Qualitative Results}
We show a few qualitative results in Fig \ref{fig:qualitative} using a detection threshold of 0.45 for images from the COCO-validation set. The R-FCN detector was used to generate detections. It is interesting to observe that Soft-NMS helps in cases when bad detections (false positives) have a small overlap with a good detection (true positive) and also when they have a low overlap with a good detection. For example, in the street image (No.8), a large wide bounding box spanning multiple people is suppressed because it had a small overlap with multiple detection boxes with a higher score than it. Hence, its score was reduced multiple times because of which it was suppressed. We observe a similar behaviour in image No.9. In the beach image (No.1), the score for the larger bounding box near the woman's handbag is suppressed below 0.45. We also see that a false positive near the bowl in the kitchen image (No.4) is suppressed. In other cases, like for zebra, horse and giraffe images (images 2,5,7 and 13), the detection boxes get suppressed with NMS while Soft-NMS assigns a slightly lower score for neighboring boxes because of which we are able to detect true positives above a detection threshold of 0.45.

% COCO | R-FCN : 0.31809(full) ; 52857(0.5) ; 33932(0.75)  | Faster : 0.25439 ; 0.46599 ; 0.24997
% VOC  | R-FCN : 51.5087 ; 0.8003; 0.5658 | Faster : 39.1959 ; 0.7121 ; 0.3811

%baseline :
% VOC  | R-FCN : 51.5087 ; 0.8003; 0.5658 | Faster : 39.1959 ; 0.7121 ; 0.3811

% \begin{table*}[t]
% \begin{center}
% %\small
% \begin{tabular}{|c|c|c|c|c|}
%   \hline
%   Method & Testing data & AP 0.5:0.95  & AP @ 0.5 & AP @0.75 \\
%   \hline\hline
%   R-FCN + L-NMS  & VOC 07 test & 51.50 & 80.03 & 56.58 \\
%   F-RCNN + L-NMS  & VOC 07 test & 39.20 & 71.21 & 38.11 \\
%   \hline
%   R-FCN + L-NMS  & ms-coco 14 minival & 31.80 & 52.86 & 33.93 \\
%   F-RCNN + L-NMS & ms-coco 14 minival & 25.43 & 45.56 & 25.00 \\
%   \hline
%  \end{tabular}
% \caption{Results on VOC 2007 and minival using Linear-NMS. }
% \label{tab:l-nms}
% \end{center}
% \end{table*}

{\small
\bibliographystyle{ieee}
\bibliography{egbib}

\begin{thebibliography}{10}\itemsep=-1pt

\bibitem{canny1986computational}
J.~Canny.
\newblock A computational approach to edge detection.
\newblock {\em IEEE Transactions on pattern analysis and machine intelligence},
  (6):679--698, 1986.

\bibitem{collins2000system}
R.~T. Collins, A.~J. Lipton, T.~Kanade, H.~Fujiyoshi, D.~Duggins, Y.~Tsin,
  D.~Tolliver, N.~Enomoto, O.~Hasegawa, P.~Burt, et~al.
\newblock A system for video surveillance and monitoring.
\newblock 2000.

\bibitem{dai2017deformable}
J.~Dai, H.~Qi, Y.~Xiong, Y.~Li, G.~Zhang, H.~Hu, and Y.~Wei.
\newblock Deformable convolutional networks.
\newblock {\em arXiv preprint arXiv:1703.06211}, 2017.

\bibitem{dalal2005histograms}
N.~Dalal and B.~Triggs.
\newblock Histograms of oriented gradients for human detection.
\newblock In {\em Computer Vision and Pattern Recognition, 2005. CVPR 2005.
  IEEE Computer Society Conference on}, volume~1, pages 886--893. IEEE, 2005.

\bibitem{desai2011discriminative}
C.~Desai, D.~Ramanan, and C.~C. Fowlkes.
\newblock Discriminative models for multi-class object layout.
\newblock {\em International journal of computer vision}, 95(1):1--12, 2011.

\bibitem{dollar2009pedestrian}
P.~Doll{\'a}r, C.~Wojek, B.~Schiele, and P.~Perona.
\newblock Pedestrian detection: A benchmark.
\newblock In {\em Computer Vision and Pattern Recognition, 2009. CVPR 2009.
  IEEE Conference on}, pages 304--311. IEEE, 2009.

\bibitem{everingham2010pascal}
M.~Everingham, L.~Van~Gool, C.~K. Williams, J.~Winn, and A.~Zisserman.
\newblock The pascal visual object classes (voc) challenge.
\newblock {\em International journal of computer vision}, 88(2):303--338, 2010.

\bibitem{felzenszwalb2010object}
P.~F. Felzenszwalb, R.~B. Girshick, D.~McAllester, and D.~Ramanan.
\newblock Object detection with discriminatively trained part-based models.
\newblock {\em IEEE transactions on pattern analysis and machine intelligence},
  32(9):1627--1645, 2010.

\bibitem{geiger2013vision}
A.~Geiger, P.~Lenz, C.~Stiller, and R.~Urtasun.
\newblock Vision meets robotics: The kitti dataset.
\newblock {\em The International Journal of Robotics Research},
  32(11):1231--1237, 2013.

\bibitem{girshick2014rich}
R.~Girshick, J.~Donahue, T.~Darrell, and J.~Malik.
\newblock Rich feature hierarchies for accurate object detection and semantic
  segmentation.
\newblock In {\em Proceedings of the IEEE conference on computer vision and
  pattern recognition}, pages 580--587, 2014.

\bibitem{haritaoglu2000w}
I.~Haritaoglu, D.~Harwood, and L.~S. Davis.
\newblock W/sup 4: real-time surveillance of people and their activities.
\newblock {\em IEEE Transactions on pattern analysis and machine intelligence},
  22(8):809--830, 2000.

\bibitem{harris1988combined}
C.~Harris and M.~Stephens.
\newblock A combined corner and edge detector.
\newblock In {\em Alvey vision conference}, volume~15, pages 10--5244.
  Citeseer, 1988.

\bibitem{he2016deep}
K.~He, X.~Zhang, S.~Ren, and J.~Sun.
\newblock Deep residual learning for image recognition.
\newblock In {\em Proceedings of the IEEE Conference on Computer Vision and
  Pattern Recognition}, pages 770--778, 2016.

\bibitem{jia2014caffe}
Y.~Jia, E.~Shelhamer, J.~Donahue, S.~Karayev, J.~Long, R.~Girshick,
  S.~Guadarrama, and T.~Darrell.
\newblock Caffe: Convolutional architecture for fast feature embedding.
\newblock In {\em Proceedings of the 22nd ACM international conference on
  Multimedia}, pages 675--678. ACM, 2014.

\bibitem{lee2016individualness}
D.~Lee, G.~Cha, M.-H. Yang, and S.~Oh.
\newblock Individualness and determinantal point processes for pedestrian
  detection.
\newblock In {\em European Conference on Computer Vision}, pages 330--346.
  Springer, 2016.

\bibitem{li2016r}
Y.~Li, K.~He, J.~Sun, et~al.
\newblock R-fcn: Object detection via region-based fully convolutional
  networks.
\newblock In {\em Advances in Neural Information Processing Systems}, pages
  379--387, 2016.

\bibitem{lin2014microsoft}
T.-Y. Lin, M.~Maire, S.~Belongie, J.~Hays, P.~Perona, D.~Ramanan,
  P.~Doll{\'a}r, and C.~L. Zitnick.
\newblock Microsoft coco: Common objects in context.
\newblock In {\em European Conference on Computer Vision}, pages 740--755.
  Springer, 2014.

\bibitem{liu2016ssd}
W.~Liu, D.~Anguelov, D.~Erhan, C.~Szegedy, S.~Reed, C.-Y. Fu, and A.~C. Berg.
\newblock Ssd: Single shot multibox detector.
\newblock In {\em European Conference on Computer Vision}, pages 21--37.
  Springer, 2016.

\bibitem{lowe2004distinctive}
D.~G. Lowe.
\newblock Distinctive image features from scale-invariant keypoints.
\newblock {\em International journal of computer vision}, 60(2):91--110, 2004.

\bibitem{mikolajczyk2004scale}
K.~Mikolajczyk and C.~Schmid.
\newblock Scale \& affine invariant interest point detectors.
\newblock {\em International journal of computer vision}, 60(1):63--86, 2004.

\bibitem{mrowca2015spatial}
D.~Mrowca, M.~Rohrbach, J.~Hoffman, R.~Hu, K.~Saenko, and T.~Darrell.
\newblock Spatial semantic regularisation for large scale object detection.
\newblock In {\em Proceedings of the IEEE international conference on computer
  vision}, pages 2003--2011, 2015.

\bibitem{philbin2007object}
J.~Philbin, O.~Chum, M.~Isard, J.~Sivic, and A.~Zisserman.
\newblock Object retrieval with large vocabularies and fast spatial matching.
\newblock In {\em Computer Vision and Pattern Recognition, 2007. CVPR'07. IEEE
  Conference on}, pages 1--8. IEEE, 2007.

\bibitem{redmon2016you}
J.~Redmon, S.~Divvala, R.~Girshick, and A.~Farhadi.
\newblock You only look once: Unified, real-time object detection.
\newblock In {\em Proceedings of the IEEE Conference on Computer Vision and
  Pattern Recognition}, pages 779--788, 2016.

\bibitem{ren2015faster}
S.~Ren, K.~He, R.~Girshick, and J.~Sun.
\newblock Faster r-cnn: Towards real-time object detection with region proposal
  networks.
\newblock In {\em Advances in neural information processing systems}, pages
  91--99, 2015.

\bibitem{rosenfeld1971edge}
A.~Rosenfeld and M.~Thurston.
\newblock Edge and curve detection for visual scene analysis.
\newblock {\em IEEE Transactions on computers}, 100(5):562--569, 1971.

\bibitem{rothe2014non}
R.~Rothe, M.~Guillaumin, and L.~Van~Gool.
\newblock Non-maximum suppression for object detection by passing messages
  between windows.
\newblock In {\em Asian Conference on Computer Vision}, pages 290--306.
  Springer, 2014.

\bibitem{rujikietgumjorn2013optimized}
S.~Rujikietgumjorn and R.~T. Collins.
\newblock Optimized pedestrian detection for multiple and occluded people.
\newblock In {\em Proceedings of the IEEE Conference on Computer Vision and
  Pattern Recognition}, pages 3690--3697, 2013.

\bibitem{sivic2003video}
J.~Sivic, A.~Zisserman, et~al.
\newblock Video google: A text retrieval approach to object matching in videos.
\newblock In {\em iccv}, volume~2, pages 1470--1477, 2003.

\bibitem{viola2001rapid}
P.~Viola and M.~Jones.
\newblock Rapid object detection using a boosted cascade of simple features.
\newblock In {\em Computer Vision and Pattern Recognition, 2001. CVPR 2001.
  Proceedings of the 2001 IEEE Computer Society Conference on}, volume~1, pages
  I--I. IEEE, 2001.

\bibitem{zhang2016unconstrained}
J.~Zhang, S.~Sclaroff, Z.~Lin, X.~Shen, B.~Price, and R.~Mech.
\newblock Unconstrained salient object detection via proposal subset
  optimization.
\newblock In {\em Proceedings of the IEEE Conference on Computer Vision and
  Pattern Recognition}, pages 5733--5742, 2016.

\bibitem{zitnick2014edge}
C.~L. Zitnick and P.~Doll{\'a}r.
\newblock Edge boxes: Locating object proposals from edges.
\newblock In {\em European Conference on Computer Vision}, pages 391--405.
  Springer, 2014.

\end{thebibliography}
}

\end{document}